\documentclass[]{bytedance_seed}

\usepackage[toc,page,header]{appendix}

\usepackage{minitoc}
\usepackage{type1cm}
\RequirePackage[T1]{fontenc}
\usepackage{amssymb}
\usepackage{pifont} % 引入 pifont 宏包
\usepackage{algorithm} 
\usepackage{algpseudocode} 
\usepackage{makecell}
\usepackage[table]{xcolor}
\usepackage{wrapfig}
\usepackage{multirow}
\usepackage{colortbl}
\usepackage{makecell}
\usepackage{soul}

\newcommand{\Ex}{\mathbb{E}}
\newcommand{\KL}{\mathbb{D}_{\text{KL}}}
\newcommand{\cmark}{\ding{51}} % 对勾
\newcommand{\xmark}{\ding{55}} % 叉号

\newcommand{\eg}{\textit{e.g.}}
\newcommand{\ie}{\textit{i.e.}}

\definecolor{lightgray}{HTML}{E8E8E8}
\definecolor{lightgreen}{HTML}{D2F3D7}
\definecolor{lightred}{HTML}{F9D1D4}

\usepackage{listings}
\usepackage{color}

\newcommand{\zeros}{\textbf{0}}

\newcommand{\bi}{{\mathbf{i}}}

\newcommand{\bx}{{\mathbf{x}}}

\newcommand{\bz}{{\mathbf{z}}}

\newcommand{\bI}{{\mathbf{I}}}

\newcommand{\balpha}{\boldsymbol{\alpha}}
\newcommand{\bbeta}{\boldsymbol{\beta}}

\newcommand{\bmu}{\boldsymbol{\mu}}
\newcommand{\btheta}{\boldsymbol{\theta}}

\newcommand{\bpsi}{\boldsymbol{\psi}}

\newcommand{\bsigma}{\boldsymbol{\sigma}}
\newcommand{\bomega}{\boldsymbol{\omega}}

\newcommand{\bepsilon}{{\boldsymbol{\epsilon}}}

\newcommand{\figref}[1]{Figure~\ref{#1}}

\newcommand{\secref}[1]{Section~\ref{#1}}

\newcommand{\tabref}[1]{Table~\ref{#1}}
\usepackage{mdframed}
\definecolor{mdproofbg}{rgb}{0.95,0.95,0.95}
{\begin{mdframed}[backgroundcolor=mdproofbg,linewidth=0]\begin{proof}}%
{\end{proof}\end{mdframed}}

\definecolor{mdworkingbg}{rgb}{1.0,0.95,0.95}
{\begin{mdframed}[backgroundcolor=mdworkingbg,linewidth=0]\begin{minipage}{\columnwidth}}%
{\end{minipage}\end{mdframed}}

\usepackage{xpatch}
\makeatletter
\AtBeginDocument{\xpatchcmd{\@thm}{\thm@headpunct{.}}{\thm@headpunct{:}}{}{}} % Use colon instead of full stop
\AtBeginDocument{\xpatchcmd{\@thm}{\fontseries\mddefault\upshape}{}{}{}} % Use bold font for theorem name
\makeatother

\definecolor{bgCode}{rgb}{0.98, 0.98, 0.98}
\definecolor{codegray}{rgb}{0.5,0.5,0.5}
\lstnewenvironment{pycode}%
{  \noindent
	\minipage{\linewidth} 
	\bigskip 
	\lstset{language=Python,numbers=left,numbersep=5pt}
	\raggedright
	\scriptsize{\textsf{Python Code}\vspace{-1.75ex}}
	\raggedleft}%
{\smallskip \endminipage}

\title{SimFlow: Simplified and End-to-End Training of \\Latent Normalizing Flows}
\author[1,2]{Qinyu Zhao}
\author[2]{Guangting Zheng}
\author[2]{Tao Yang}
\author[2\dagger]{Rui Zhu}
\author[1]{Xingjian Leng}
\author[1]{\\Stephen Gould}
\author[1]{Liang Zheng}

\affiliation[1]{Australian National University}
\affiliation[2]{ByteDance Seed}
\contribution[\dagger]{Project lead}

\abstract{
Normalizing Flows (NFs) learn invertible mappings between the data and a Gaussian distribution. Prior works usually suffer from two limitations. First, they add random noise to training samples or VAE latents as data augmentation, introducing complex pipelines including extra noising and denoising steps. Second, they use a pretrained and frozen VAE encoder, resulting in suboptimal reconstruction and generation quality. In this paper, we find that the two issues can be solved in a very simple way: just fixing the variance (which would otherwise be predicted by the VAE encoder) to a constant (\eg, 0.5). On the one hand, this method allows the encoder to output a broader distribution of tokens and the decoder to learn to reconstruct clean images from the augmented token distribution, avoiding additional noise or denoising design. On the other hand, fixed variance simplifies the VAE evidence lower bound, making it stable to train an NF with a VAE jointly. On the ImageNet $256 \times 256$ generation task, our model SimFlow obtains a gFID score of 2.15, outperforming the state-of-the-art method STARFlow (gFID 2.40). Moreover, SimFlow can be seamlessly integrated with the end-to-end representation alignment (REPA-E) method and achieves an improved gFID of 1.91, setting a new state of the art among NFs. 
}

\date{\today}
\correspondence{\email{qinyu.zhao@anu.edu.au}}

\checkdata[Project Page]{\url{https://qinyu-allen-zhao.github.io/SimFlow/}}

\begin{document}
\maketitle

%不需要目录就注释掉 注意目录不要和第一页放在一块 要有\newpage
%\newpage
%\tableofcontents
%\newpage

\section{Introduction}\label{sec:intro}
Normalizing flows (NFs)~\citep{vnf,nvp,glow,tarflow,starflow,tarflowllm,starflow_v} model a data distribution by transforming a prior distribution (\eg, the normal distribution) via invertible mappings. For easier training and better generation, state-of-the-art methods \citep{starflow,starflow_v} adopt two strategies: 1) using a variational autoencoder (VAE)~\citep{sd} to train a latent NF, and 2) adding noise to VAE latents during NF training as data augmentation. 

However, there are two limitations with the above strategies. First, while adding random noise smooths latent space for more generalizable NF modeling,
it complicates the pipelines. Specifically, NFs trained with noisy inputs generate noisy outputs, so existing works have to introduce an additional denoising process, such as score-based denoising~\citep{tarflow}, a fine-tuned VAE decoder~\citep{starflow}, or a flow matching model for denoising~\citep{starflow_v}. The extra noising and denoising steps complicate the training and inference processes.

Second and more importantly, these methods use a frozen VAE encoder, which leads to suboptimal reconstruction and generation quality. For \textit{reconstruction}, the encoder is sensitive to noise because it is not trained under noise augmentation introduced for NFs. Possibly, if images are very close to each other in the latent space, noise perturbations may severely degrade the latents. While it is feasible to fine-tune the decoder~\citep{starflow}, reconstruction quality remains poor, because some encoded information is already lost due to noise (for results see Section~\ref{sec:main_eval}). For \textit{generation}, now that the encoder is not optimized jointly with an NF, the latent space may not be best suitable for NF modeling.

\begin{figure*}
%\vskip -0.15in
  \centering
  \includegraphics[width=0.98\textwidth,trim=0 290 20 0,clip]{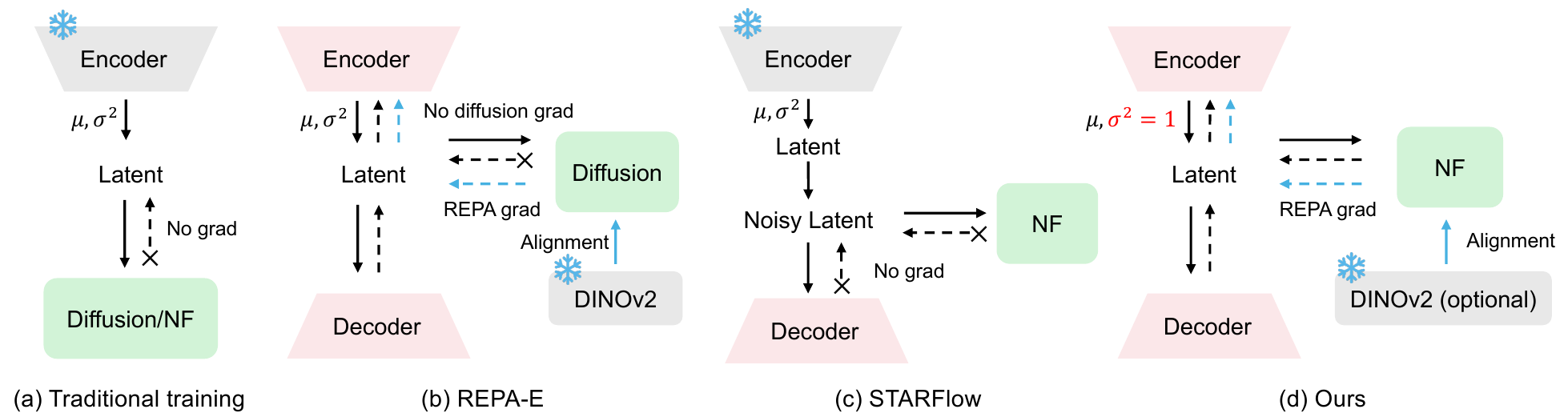}
  %\vskip -0.15in
  \caption{\textbf{Comparison of our framework with closely-related methods.} (a) Standard practice trains a VAE first and then train a generative model with the VAE frozen. Note that, for each image, the VAE encoder outputs the mean $\bmu$ and the variance $\bsigma^2$ of a Gaussian, from which a set of tokens is sampled. The variance $\bsigma^2$ is usually very small in a standard pretrained VAE. (b) REPA-E~\citep{repa_e} jointly trains diffusion with VAE using the REPA loss \cite{repa}, and the diffusion gradient is stopped before the VAE to avoid latent collapse (where the token variation decreases and the generation quality degrades). (c) STARFlow~\citep{starflow} trains NF and decoder on noisy latent with the encoder frozen. (d) We train an NF and a VAE in an end-to-end way from scratch. There is no stop-gradient operator, significantly simplifying prior frameworks. Solid arrows indicate the forward pass, while dashed arrows denote gradient flows. We label frozen modules in \sethlcolor{lightgray}\hl{gray}, 
generative models in \sethlcolor{lightgreen}\hl{green}, 
and VAE modules involved in training in \sethlcolor{lightred}\hl{red}.}
  \label{fig:framework}
% \vskip -0.2in
\end{figure*}

\begin{wrapfigure}{r}{0.5\linewidth}
    \centering
    \vspace{-16pt} % adjust vertical spacing if needed
  \includegraphics[width=0.46\textwidth,trim=10 10 10 0,clip]{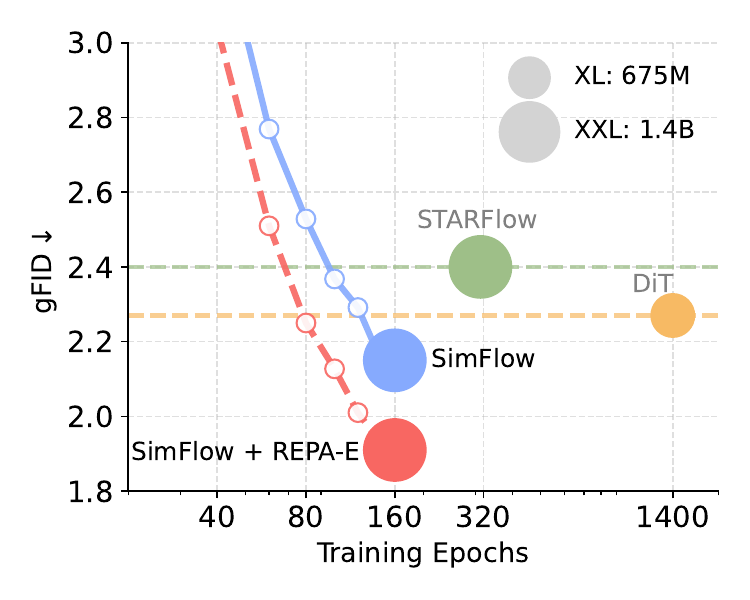}
  %\vskip -0.15in
  \caption{\textbf{Comparing SimFlow with prior works.} On ImageNet $256\times256$, our end-to-end trained model SimFlow achieves significantly better generation
quality than the state-of-the-art NF model STARFlow \cite{starflow} with much fewer training epochs. Training SimFlow with REPA-E \cite{repa_e} further improves gFID.}
  \label{fig:figure1}
\end{wrapfigure}

In this paper, we solve both issues at the same time by introducing SimFlow, a simple end-to-end training framework for latent NFs. Our key idea is to fix the encoder-predicted variance to a constant (for instance, $0.5$). On the one hand, this method has similar effects with adding noise to latents~\citep{starflow}, that is, to smooth the latent space and improve generalization, but greatly simplifies the pipeline. Note that pretrained VAEs tend to predict extremely small variance (\eg, less than 0.01), which collapses a token distribution to nearly a single point. In contrast, our method of fixing variance to a larger value ensures each latent token is sampled from a broader distribution rather than a nearly deterministic result. Meanwhile, the decoder learns how to reconstruct clean images from the augmented token distributions during VAE training. In this way, our approach removes the need for additional noising or denoising design. 

On the other hand, fixing the variance to a value like 0.5 enables effective end-to-end training of latent NFs, making the VAE latents compatible with NF modeling. From the VAE evidence lower bound (ELBO) perspective, this method makes the entropy term become a constant, which simplifies the training objective to consist of only the reconstruction loss and the generation loss. We find that it is much easier to balance between reconstruction and generation than to train the whole framework. 
A comparison of SimFlow with prior works is shown in \figref{fig:framework}.

On the ImageNet $256\times256$ generation benchmark, SimFlow achieves a gFID score of 2.15, surpassing the prior work STARFlow (gFID 2.40)~\citep{starflow}, as shown in \figref{fig:figure1}. This is also the first report that NFs outperform DiT~\citep{dit}, a representative diffusion model. While SimFlow has a larger model size, its convergence rate is more than 8 times faster than DiT. Importantly, our framework is fully compatible with REPA-E~\citep{repa_e}: aligning NF features with DINOv2~\citep{dino} and training both the VAE and NF using the alignment loss. Combining SimFlow with REPA-E further decreases gFID to 1.91 on Image $256\times256$ and achieves a gFID score of 2.74 on ImageNet $512\times512$, establishing new state-of-the-art performance among NF-based models.

\section{Related work}\label{sec:related_works}
\textbf{Normalizing flows (NFs)}~\cite{vnf,papamakarios2021normalizing,kobyzev2020normalizing,nice,nvp,glow,van2016pixel,draxler2024free,mathieu2020riemannian,giaquinto2020gradient,draxler2024universality,mate2022flowification,tarflowllm} are a useful framework for density estimation, visual generation, and text generation. In this paper, we mainly adopt autoregressive flows (AFs)~\citep{tarflow,starflow}. In each invertible transformation of an AF, each token is transformed conditioned on previous tokens. Representative AFs include IAF~\cite{iaf}, MAF~\cite{maf}, neural AF~\cite{huang2018neural}, and T-NAF~\cite{t_naf}. More recently, TARFlow~\cite{tarflow} leverages causal Transformers and simplifies the log-determinant term in the loss function, leading to notable improvements in generation quality. STARFlow~\citep{starflow} extends TARFlow into the VAE latent space and further improves the NF performance.

\textbf{VAEs with fixed variance or noise-augmented training.} \citet{sun2024multimodal} introduce $\sigma$-VAE, which has a fixed variance, and later works adopted this technique \citep{team2025nextstep,ke2025hyperspherical}. Other studies~\citep{yang2025latent,qiu2025image,meshchaninov2025compressed} introduce noise or perturbation in the latent space. While these approaches are similar to ours, they are motivated differently and do not explore joint training with generative models. In Section~\ref{sec:working_mech}, we show that our perspective provides a unified explanation of why these methods actually allow for stable end-to-end training.

\textbf{Joint training of generative models with VAEs.} The standard practice is to train a VAE on the reconstruction task first and then train a generative model with the VAE frozen~\citep{dit,sit,mar,ren2024flowar,xar,pang2025randar,llamagen,starflow,bao2023all,gu2024kaleido,gu2024dart,shen2024many}. But an important question is whether the pretrained VAE space is suitable for training a generative model. Without solving this, a possible consequence is that a VAE with excellent reconstruction ability leads to poor generation quality of the generative models trained on the latent space~\citep{vavae,eqvae}. 

End-to-end training of generative models with VAEs is appealing because it not only streamlines the training pipeline but  is also expected to make the VAE latent space more suitable for generation. Early exploration~\citep{lsgm,vnf,fvae} is not stable in training and does not achieve competitive performance on a large-scale dataset like ImageNet $256\times 256$. Recently, REPA-E~\citep{repa_e} reported an FID of 1.12 on ImageNet $256\times 256$ with end-to-end training. In REPA-E, the gradient of the REPA loss~\citep{repa} flows through the diffusion model to the VAE, while the gradient from the diffusion model does not flow to the VAE (see \figref{fig:framework}(b)). When the latter is allowed, they observe latent space collapse, where the latent variance shrinks quickly, and the generation quality is poor. %, if they do not stop the diffusion gradient to VAE. 
Different from REPA-E, SimFlow jointly trains NFs and VAEs from scratch without stopping any gradient, further simplifying the training framework.

\section{Preliminaries}\label{sec:preliminaries}
A standard VAE~\citep{kingma2013auto} consist of an encoder $q_{\bpsi}$ and a decoder $p_{\bomega}$. Given an image $\bi$, the encoder predicts mean $\bmu$ and variance $\bsigma^2$ of a Gaussian distribution $\mathcal{N}(\bmu, \text{diag}(\bsigma^2))$. A latent variable $\bx$ (a set of tokens) is sampled from $\mathcal{N}(\bmu, \text{diag}(\bsigma^2))$. The decoder is trained to reconstruct the image $\bi$ from $\bx$. The VAE is trained with ELBO as follows:
\begin{equation}
    \log p(\bi) \ge \mathbb{E}_{\bx\sim q_{\bpsi}(\bx \mid \bi)} \left[ \underbrace{\log p_{\bomega}(\bi \mid \bx)}_{\text{Reconstruction}} + \underbrace{\log p(\bx)}_{\text{Prior}} - \underbrace{\log q_{\bpsi}(\bx \mid \bi)}_{\text{Entropy}}\right],
\end{equation}
where the prior term is usually chosen as the normal distribution $\mathcal{N}(\zeros, \bI)$, and the prior and entropy terms are combined into a Kullback-Leibler (KL) divergence term.

Latent NFs~\cite{vnf,nice,nvp,glow,starflow} map the VAE latent distribution into a simple one $\bz \sim p_0(\bz)$ (the normal distribution), via learning an invertible function $f_{\btheta}$. During training, the forward pass maps the sampled latent $\bx$ to $\bz=f_{\btheta}(\bx)$, following the change of variable formula:
\begin{equation}
p_{\text{NF}}(\bx;{\btheta}) = p_0(\bz)\left| \det \left( \frac{\partial \bz}{\partial \bx} \right) \right|=p_0(f_{\btheta}(\bx)) \left| \det \left( \frac{\partial f_{\btheta}(\bx)}{\partial \bx} \right) \right|.\label{eq:change_of_var}
\end{equation}
NFs are trained with maximum likelihood estimation:
\begin{equation}
\max_{\btheta}\mathbb{E}_{\bx \sim p_{\text{latent}}} \log p_{\text{NF}}(\bx; {\btheta}).
\end{equation}

At inference, $\bz$ is sampled from the target distribution $\bz\sim \mathcal{N}(\zeros, \bI)$, and an inverse process maps $\bz$ to $\bx=f_{\btheta}^{-1}(\bz)$, which is then decoded by the decoder to images. In this way, $f_{\btheta}^{-1}$ with the VAE decoder is a generative model, which maps Gaussian noise to new image samples.

This study mainly builds on STARFlow~\citep{starflow}, as it achieves competitive performance compared to other generative models on the large-scale ImageNet dataset. That said, our end-to-end training framework is expected to be applicable to other latent NF models as well.

\section{Method}\label{sec:method}
In Section~\ref{sec:simflow}, we present SimFlow, our joint training framework for latent NFs, from two different perspectives: noise augmentation in NF training and the VAE ELBO formulation. In Section \ref{sec:repa}, we adopt REPA-E to further improve SimFlow. In Section \ref{sec:working_mech}, latent space is analyzed to show the effects of end-to-end training. Section \ref{sec:cfg} revisits and improves the classifier-free guidance (CFG) for NFs.

\subsection{SimFlow: End-to-end training of latent NFs}\label{sec:simflow}
Let $f_{\btheta}$ denote the NF model, and let $q_{\bpsi}$ and $p_{\bomega}$ denote the encoder and decoder of a VAE, respectively. We set variance outputted from the encoder as a constant $\bar{\sigma}^2$. The encoder only predicts mean $\bmu$ for each image $\bi$, and the latent variable $\bx$ is sampled from the Gaussian $\mathcal{N}(\bmu, \bar{\sigma}^2\bI)$. The NF is trained to convert $\bx$ to Gaussian samples $\bz=f_{\btheta}(\bx)$, and the decoder learns to reconstruct the images from $\bx$.

Our system is trained with the combined VAE and NF objective (details are in \secref{sec:supp_training_obj}):
\begin{equation}
\max_{\btheta,\bpsi,\bomega}\mathbb{E}_{\bi \sim p_{\text{data}}} \mathbb{E}_{\bx\sim q_{\bpsi}(\bx \mid \bi)} \left[ \log p_{\bomega}(\bi \mid \bx) + \log p_{\text{NF}}(\bx; {\btheta}) \right],
\label{eq:simflow_train}
\end{equation}
where the entropy term is a constant and thus omitted. Note that we do not tune the weights of loss terms in Eq.~\ref{eq:simflow_train} and simply set them to 1.0. We also add perceptual loss and adversarial loss~\citep{gan} for VAE training~\citep{sd}.

\begin{figure*}
%\vskip -0.15in
  \centering
  \includegraphics[width=0.98\textwidth,trim=0 140 10 0,clip]{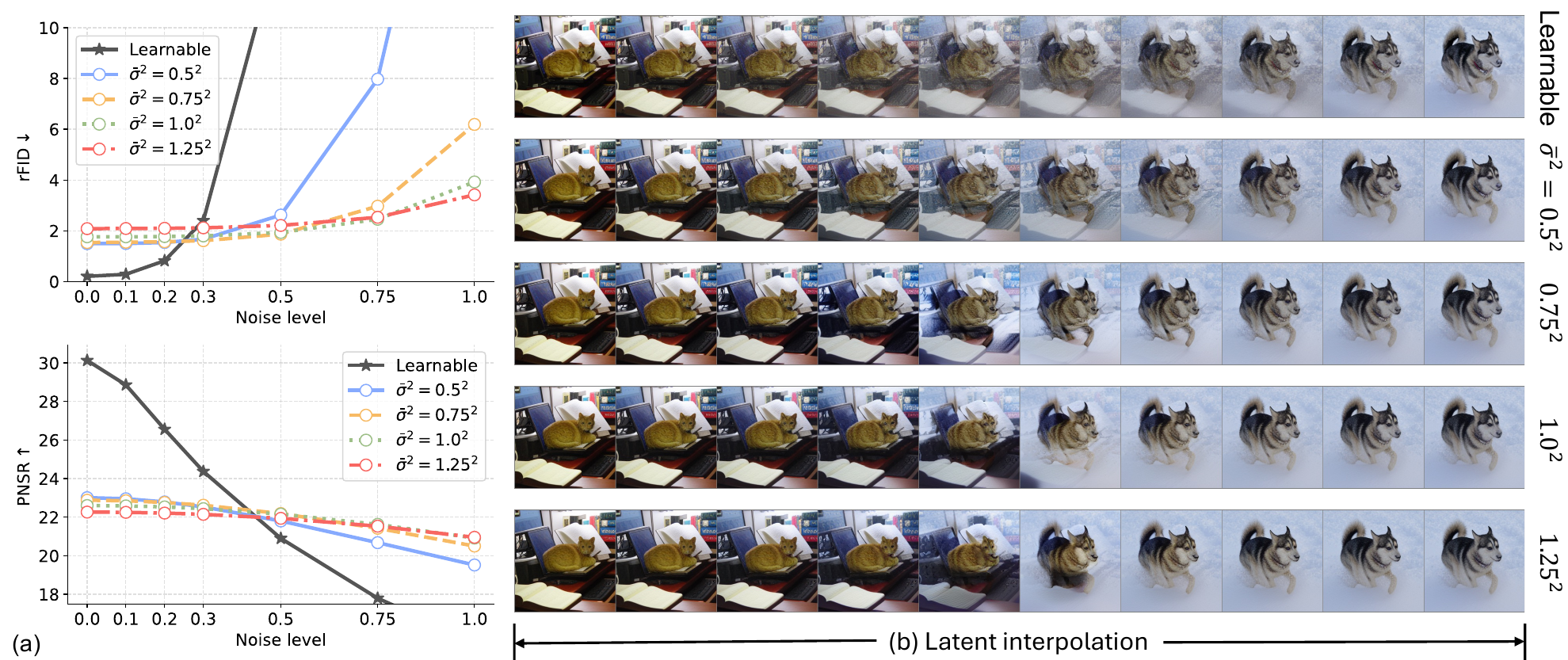}
  %\vskip -0.15in
  \caption{\textbf{Robustness of VAEs with fixed variances.} (a) A VAE with a large and fixed variance can maintain reconstruction quality under latent noise, while the performance of a VAE with learnable variance degrades significantly. (b) For VAEs with a large variance, the images reconstructed from linearly interpolated latents still clearly show the main subjects (the cat or the dog), rather than blending them. `Learnable' indicates a standard VAE with learnable variance, while `$\bar{\sigma}^2=x^2$' denotes a VAE with a fixed variance of $x^2$.}
  \label{fig:vae_robustness}
% \vskip -0.2in
\end{figure*}
% At inference time, points $\bz$ are sampled from Gaussian distribution, and we run the inverse process of NFs to get $\bx = f^{-1}_{\btheta}(\bz)$. Images are decoded by $p_{\bomega}$ based on $\bx$.

\textbf{From the noise-augmented training perspective,} the distribution outputted by the VAE encoder can be decomposed into $\bmu$ and a Gaussian noise, $\bx=\bmu+\bepsilon, \bepsilon\sim\mathcal{N}(\zeros, \text{diag}(\bsigma^2))$. Ideally, the token distribution is already augmented by noise $\bepsilon$. However, for a well-pretrained VAE, we observe that $\bsigma^2$ is usually very small. There are two main reasons. First, the KL term could have maintained the variance as regularization but is usually assigned a very small weight (\textit{e.g}., $10^{-5}$), making its effect negligible. Second, if $\bsigma^2$ is large, it is hard for the decoder to reconstruct the image from the highly varying latent. Thus, the encoder shrinks the variance, making it easier to decode. As a result, the latent still needs noise augmentation for NF training.

In this work, we manually set $\sigma^2$ from the encoder to a constant $\bar{\sigma}^2$ and the VAE encoder embeds images into $\bx=\bmu+\bepsilon, \bepsilon\sim\mathcal{N}(\zeros, \bar{\sigma}^2\bI)$. The resulting `VAE+NF' framework naturally becomes noise-augmented training. NF learns the distribution while the decoder is trained to reconstruct the image. This avoids additional noising and denoising steps, simplifying the prior frameworks.

\textbf{From the VAE ELBO perspective,} a straightforward way for end-to-end training is to follow the ELBO:
\begin{equation}
    \mathbb{E}_{\bx\sim q_{\bpsi}(\bx \mid \bi)} \left[ \log p_{\bomega}(\bi \mid \bx) + \log p_{\text{NF}}(\bx; {\btheta}) - \log q_{\bpsi}(\bx \mid \bi)\right],
\end{equation}
where the entropy term maintains variance not to be small, avoiding latent collapse~\citep{lsgm,repa_e}. 

However, we empirically find it hard to balance the loss terms. The reconstruction term tends to reduce the variance, the NF term tends to shrink the predicted latent mean, while the entropy term tends to enlarge the variance. The tension makes the training sensitive to the loss weights and leads to suboptimal performance.

In SimFlow, because we manually set $\bsigma^2$ as a constant, the entropy term becomes constant as well (see 
\secref{sec:supp_entropy_term}). The objective now consists of only the reconstruction and generation terms. It then becomes easier to train the entire framework by simply assigning equal weights to both terms.

\begin{figure*}
%\vskip -0.15in
  \centering
\includegraphics[width=0.98\textwidth,trim=0 265 0 0,clip]{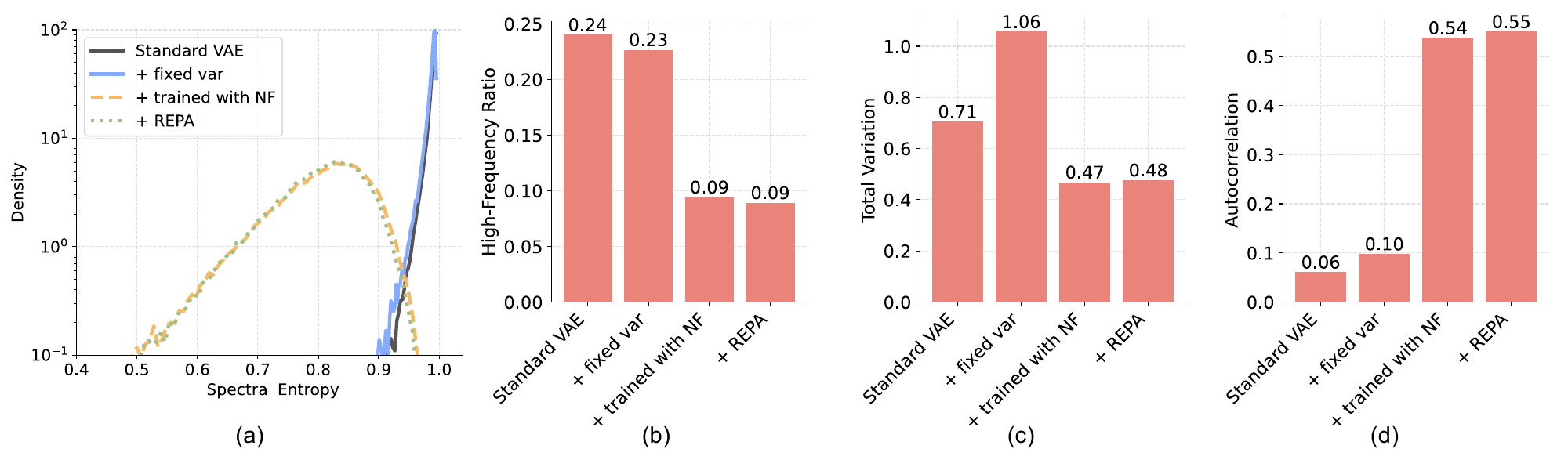}
  %\vskip -0.15in
  \caption{\textbf{End-to-end training makes latent space more suitable for developing generative models.} (a) Spectral entropy measures the randomness of frequency components; lower values indicate simpler data distributions in the frequency domain. (b) Ratio of high-frequency components. (c) Total variation captures the overall local changes across tokens; lower values imply smoother latents. (d) Autocorrelation reflects how similar a token sequence is to a shifted version of itself; higher autocorrelation indicates stronger spatial consistency. }
  \label{fig:vae_latent}
% \vskip -0.2in
\end{figure*}

\subsection{Representation alignment}
\label{sec:repa}
Recent studies~\citep{repa,vavae,repa_e} show that alignment with features extracted from a representative model significantly benefits the training of a diffusion model. During training, REPA~\citep{repa} extracts features with a pretrained representation model, such as DINOv2-B~\citep{dino} and uses a projector to align the hidden states from a diffusion block with the DINO features. Besides, VAVAE~\citep{vavae} and REPA-E~\citep{repa_e} report stronger benefits of alignment when training VAE or under joint training. 

We adopt REPA to SimFlow to further improve the reconstruction and generation quality. As shown in \figref{fig:framework}(d), we extract features from DINOv2-B and use a projector to align the hidden states of NF with DINO features. Note that, because of the end-to-end nature of SimFlow, REPA applied to SimFlow naturally becomes REPA-E, where REPA gradients flow from the NF to the VAE encoder.

\subsection{Working mechanism}\label{sec:working_mech}
\textbf{Why does a fixed variance improve NF generation?} % To answer this, we design two different analyses to show the improved robustness of VAEs with a fixed variance.
\textbf{First}, our VAE is more robust to imperfect predictions from the generative model. We use different VAEs to reconstruct the 50K validation images in ImageNet $256\times256$ with increasing noise level on the latent, and then measure the reconstruct quality. \figref{fig:vae_robustness}(a) shows that while a VAE with learnable variance achieves the best reconstruction quality when there is no noise on the latent, VAE with a large and fixed variance is more robust to noise. Thus, when NFs generate imperfect latent, our VAE can still decode good images.
\textbf{Second}, we linearly interpolate the latents from two images and visualize the reconstruction results of the interpolation points. As seen in \figref{fig:vae_robustness}(b), reconstruction results are better for the VAE with fixed variance, indicating that the latent space of our VAEs is smooth and easy for NFs to learn. Note that, while large variance makes the VAE more robust, it may over-smooth the latent space and limit the reconstruction and generation quality. Thus, a moderate variance performs overall the best (see Section~\ref{sec:further_ana}).

\textbf{Why does fixed variance avoid latent collapse?}
\citet{repa_e} observe latent space collapse when they naively train DiT and VAE together. We observe the same phenomenon on latent NFs. The key reason is that the loss of generative models encourages the latent variables 
$\bx$ of different images to be close to each other. When the latent variance becomes smaller, the MSE loss term for DiTs or NFs is reduced, providing a shortcut for optimization and resulting in the latent collapse.

As discussed in \secref{sec:simflow}, our encoder with fixed variance can be considered as a predicted mean $\bmu$ plus noise: $\bx=\bmu+\bepsilon, \bepsilon\sim\mathcal{N}(\zeros, \bar{\sigma}^2\bI)$. If $\bmu$ of different images were close to each other, the differences in $\bmu$ would be overwhelmed by noise $\bepsilon$, making it hard to reconstruct images. To maintain good reconstruction quality, the system thus tends not to shrink the latent variation as much as naive end-to-end training, allowing for effective end-to-end training. We show similar effects of other noise design in Section~\ref{sec:further_ana}.

\textbf{How does end-to-end training help?} Prior works explore what VAE latent space is suitable for training a generative model, especially for diffusion models~\citep{yang2025latent,diffusability,vavae}. Based on SimFlow, we present a new perspective: if we jointly train VAE with a generative model, will the latent space become more suitable for generation? 

To answer this, we use different statistics to analyze latent space. As shown in \figref{fig:vae_latent}, the latent space of end-to-end training has lower spectral entropy and less high-frequency components, making it easier for generative models to fit the distribution~\citep{diffusability}. Moreover, tokens have lower total variation and higher autocorrelation. This means the token sequence is more suitable for autoregressive modeling. Because our NF is based on AF, the latent space is therefore expected to be more favorable for NF modeling.

\subsection{Revisit classifier-free guidance for NFs}
\label{sec:cfg}
CFG~\citep{cfg} significantly improves the performance of diffusion models, by using unconditional outputs to push the class-conditional predictions. This technique has been applied to NFs recently~\citep{tarflow,starflow}. TARFlow~\citep{tarflow} does an early exploration but their method lacks theoretical foundation. STARFlow~\citep{starflow} carefully designs CFG for NFs based on mathematical proof, but their CFG can only be applied on the last NF block. 

Inspired by the denoising step in TARFlow~\citep{tarflow}, we run STARFlow CFG first to generate tokens and then derive a score-based step on the generated tokens, \ie,
\begin{equation}\label{eq:cfg}
\Tilde{\bx} = \bx + \gamma (\nabla_\bx \log p_{\text{NF}}(\bx|c)-\nabla_\bx \log p_{\text{NF}}(\bx|\phi)),
\end{equation}
where $\nabla_\bx \log p_{\text{NF}}(\bx|c)$ and $\nabla_\bx \log p_{\text{NF}}(\bx|\phi)$ are the gradients of the NF with and without class labels $c$, respectively. $\gamma$ controls the step size.
This step leverages the full distribution modeled by the NF and utilizes all NF blocks, rather than only the final one, thereby further improving the generation quality over the CFG method in STARFlow. We will compare our CFG with STARFlow in \secref{sec:main_eval}.

\begin{table}[t]
% \vskip -0.1in
\scriptsize 
    \centering
    \caption{\textbf{Class-conditional performance on ImageNet 256$\times$256.} Methods requiring external pretrained models (\eg, DINOv2-B) for alignment are highlighted in \sethlcolor{blue!8}\hl{blue}.}
    \setlength{\tabcolsep}{1.5mm}{
\begin{tabular}{l cccc cccc cccc}
\toprule
\multirow{2}{*}{Method} & \multirow{2}{*}{\makecell{Epochs}} & \multirow{2}{*}{\#Params} & \multicolumn{2}{c}{VAE} & \multicolumn{4}{c}{W/o guidance} & \multicolumn{4}{c}{W/ guidance} \\
\cmidrule(lr){4-5} \cmidrule(lr){6-9} \cmidrule(lr){10-13}
 & & & rFID$\downarrow$ & PSNR$\uparrow$ & gFID$\downarrow$ & IS$\uparrow$ & Prec.$\uparrow$ & Rec.$\uparrow$ & gFID$\downarrow$ & IS$\uparrow$ & Prec.$\uparrow$ & Rec.$\uparrow$ \\
\midrule
\multicolumn{11}{l}{\textit{Pixel Space}} \\
\arrayrulecolor{black!30}\midrule
ADM~\citep{adm} &  400  &  554M & - & - & 10.94 &  101.0 & 0.69 & 0.63 & 3.94 & 215.8 & 0.83 & 0.53\\
RIN~\citep{jabri2022scalable} &  480  &  410M & -  & - & 3.42  & 182.0  &  -   &  -   &  -   &   -    &  -   &  -  \\
PixelFlow~\citep{chen2025pixelflow} & 320 & 677M & - & - & - & -   &   -  &  -   & 1.98 & 282.1 & 0.81 & 0.60 \\
PixNerd~\citep{wang2025pixnerd} & 160 & 700M & - & - & -   &  -       &  - &  -  &  2.15 & 297.0 & 0.79 & 0.59 \\
SiD2~\citep{hoogeboom2024simpler} &  1280   &   -   & - & - & -   &  - &  -   &  -   &  1.38   &   -    &  -   &  - \\
TARFlow~\citep{tarflow} & 320 & 1.4B & - & - & - & - & - & - & 4.69 & - & - & - \\
JetFormer~\citep{tschannen2024jetformer} & 500 & 2.8B & - & - & - & - & - & -  & 6.64 & - & 0.69 & 0.56 \\
FARMER~\citep{zheng2025farmer} & 320 & 1.9B & - & - & - & - & - & - & 3.60 & 269.2 & 0.81 & 0.51\\
\arrayrulecolor{black}\midrule
\multicolumn{11}{l}{\textit{Latent Autoregressive}} \\
\arrayrulecolor{black!30}\midrule
VAR~\citep{VAR} &  350  &  2.0B  & - & - & 1.92 & 323.1  & 0.82 & 0.59 & 1.73 & 350.2 & 0.82 & 0.60\\
MAR~\citep{mar} &  800  &  943M  & 0.53 & 26.18 & 2.35 &  227.8 & 0.79 & 0.62 & 1.55 & 303.7 & 0.81 & 0.62\\
xAR~\citep{xar} &  800  & 1.1B  & 0.53 & 26.18 & - & - & - & - & 1.24 & 301.6 & 0.83 & 0.64\\
\arrayrulecolor{black}\midrule
\multicolumn{11}{l}{\textit{Latent Diffusion}} \\
\arrayrulecolor{black!30}\midrule
DiT~\citep{dit} & 1400 & 675M & 0.61 & 24.98 & 9.62 & 121.5 & 0.67 & 0.67 & 2.27 & 278.2 & 0.83 & 0.57 \\
MaskDiT~\citep{maskdit} & 1600 & 675M & 0.61 & 24.98 & 5.69 & 177.9 & 0.74 & 0.60 & 2.28 & 276.6 & 0.80 & 0.61 \\
SiT~\citep{sit} & 1400 & 675M & 0.61 & 24.98 & 8.61 & 131.7 & 0.68 & 0.67 & 2.06 & 270.3 & 0.82 & 0.59 \\
MDTv2~\citep{gao2023mdtv2} & 1080 & 675M & 0.61 & 24.98 & - & - & - & - & 1.58 & 314.7 & 0.79 & 0.65 \\
\rowcolor{blue!8}REPA~\citep{repa} & 800 & 675M & 0.61 & 24.98 & 5.78 & 158.3 & 0.70 & 0.68 & 1.29 & 306.3 & 0.79 & 0.64 \\
\rowcolor{blue!8}VA-VAE~\citep{vavae} & 800 & 675M & 0.28 & 26.32 & 2.17 & 205.6 & 0.77 & 0.65 & 1.35 & 295.3 & 0.79 & 0.65 \\
\rowcolor{blue!8}DDT~\citep{decoupled_dit} & 400 & 675M & 0.61 & 24.98 & 6.27 & 154.7 & 0.68 & 0.69 & 1.26 & 310.6 & 0.79 & 0.65\\
\rowcolor{blue!8}REPA-E~\citep{repa_e}  & 800 & 675M & 0.28 & 26.25 & 1.69 & 219.3 & 0.77 & 0.67 & 1.12 & 302.9 & 0.79 & 0.66 \\
\rowcolor{blue!8}RAE~\citep{rae} & 800 & 839M & 0.57 & 18.86 & 1.51 & 242.9 & 0.79 & 0.63 & 1.13 & 262.6 & 0.78 & 0.67 \\
 \arrayrulecolor{black}\midrule
\multicolumn{11}{l}{\textit{Latent Normalizing Flows}} \\
\arrayrulecolor{black!30}\midrule
STARFlow~\citep{starflow} & 320 & 1.4B & 2.73 & - & - & - & - & - & 2.40 & - & - & - \\
SimFlow & 160 & 1.4B & 1.21 & 23.07 & 13.72 & 105.2 & 0.67 & 0.62 & 2.15 & 276.8 & 0.83 & 0.57 \\
\rowcolor{blue!8}SimFlow + REPA-E & 160 & 1.4B & 1.08 & 23.17 & 10.13 & 124.7 & 0.71 & 0.61 & 1.91 & 284.4 & 0.82 & 0.60 \\
\arrayrulecolor{black}\bottomrule
\end{tabular}
}
\label{tab:imagenet256_sota}
% \vspace{-0.7cm}
\end{table}

\begin{table}[t]
% \vskip -0.1in
\scriptsize 
    \centering
    \caption{\textbf{Class-conditional performance on ImageNet 512$\times$512.} SimFlow with REPA-E achieves better performance than DiT and STARFlow on generating higher-resolution images.}
    \setlength{\tabcolsep}{2.0mm}{
\begin{tabular}{lccccc}
\toprule
Method & \#Params & gFID$\downarrow$  &  IS$\uparrow$  & Prec.$\uparrow$ & Rec.$\uparrow$ \\
\midrule
BigGAN-deep~\citep{biggan} & 158M & 8.43 & 177.9 & 0.88 & 0.29 \\
StyleGAN-XL~\citep{sauer2022stylegan} & -  & 2.41 &  267.8 & 0.77 & 0.52 \\
\arrayrulecolor{black}\midrule
VAR~\citep{VAR} & 2.3B & 2.63 & 303.2 & - & - \\
MAGVIT-v2~\citep{magvitv2} & 307M & 1.91 & 324.3 & - & - \\
MAR~\citep{mar} & 481M & 1.73 & 279.9 & - & - \\
XAR~\citep{xar} & 608M & 1.70 & 281.5 & - & - \\
\arrayrulecolor{black}\midrule
ADM~\citep{adm} & 731M & 3.85 &  221.7 & 0.84 & 0.53\\
SiD2  & - & 1.50   &    -     &   -  &   -  \\
\arrayrulecolor{black!30}\midrule
DiT~\citep{dit} & 674M & {3.04} &  {240.8} & 0.84 & 0.54 \\
SiT~\citep{sit} & 674M & 2.62 &   252.2 & 0.84 & 0.57 \\
DiffiT~\citep{diffit}  & - & 2.67  &  252.1 & 0.83 & 0.55 \\
REPA~\citep{repa} & 675M & 2.08 &  274.6 & 0.83 & 0.58 \\
DDT~\citep{decoupled_dit} & 675M & 1.28 &  305.1 & 0.80 & 0.63 \\
EDM2~\citep{edm2} & 1.5B &1.25 & -& -& -\\
RAE~\citep{rae} & 839M & 1.13 & 259.6 & 0.80 & 0.63 \\
\arrayrulecolor{black}\midrule
STARFlow~\citep{starflow} & 1.4B & 3.00 & -& -& - \\
SimFlow + REPA-E & 1.4B & 2.74 & 304.9 & 0.81 & 0.57 \\
\arrayrulecolor{black}\bottomrule
\end{tabular}
}
\label{tab:imagenet512_sota}
% \vspace{-0.7cm}
\end{table}

\section{Experiments}\label{sec:experiments}
\subsection{Implementation details}
Experiments are conducted on the ImageNet $256\times256$ and $512\times 512$ generation tasks~\citep{imagenet}. We report gFID~\citep{fid}, IS~\citep{is}, Precision, and Recall for generation~\citep{adm}. Besides, rFID, PSNR~\citep{psnr}, LPIPS~\citep{lpips}, and SSIM~\citep{ssim} are reported for VAE reconstruction evaluation.

We use the MAR VAE architecture~\citep{sd,mar} and TARFlow as the basic NF model. Both VAE and NF are trained from scratch. Following STARFlow, we use a deep-shallow architecture. Specifically, the NF model has six blocks with 1152 hidden dimensions. The first five blocks have two layers while the last deep block has 46 layers. Unless stated otherwise, SimFlow is trained for 80 epochs with a global batch size of 256 and a constant learning rate of $1.0\times 10^{-4}$. In \tabref{tab:imagenet256_sota}, we extend the training with extra 80 epochs to get further improved performance, where the learning rate reduces from $1.0\times 10^{-4}$ to $1.0\times 10^{-6}$ at a cosine rate like STARFlow. Details are presented in \secref{sec:supp_imple_detail}.

\subsection{Main evaluation}\label{sec:main_eval}
\textbf{Comparison with state-of-the-art methods on ImageNet 256$\times$256.} Results are shown in \tabref{tab:imagenet256_sota}. 
SimFlow achieves a gFID of 2.15, significantly outperforming the prior NF method STARFlow (gFID=2.40). Note that STARFlow is trained for 320 epochs while SimFlow is trained for 160 epochs. Our jointly-trained VAE has better reconstruction quality (rFID=1.21) than STARFlow (rFID=2.73). Besides, SimFlow surpasses a representative diffusion model, DiT~\citep{dit}. Although SimFlow needs more parameters, the convergence rate is more than 8 times faster than DiT. 

With REPA-E, SimFlow establishes the new state-of-the-art performance among NFs (gFID 1.91). REPA-E not only improves the generation quality in terms of gFID but also the reconstruction performance of VAE, with better rFID and PSNR. It indicates that the guidance from a pretrained model is helpful for building a better latent space, benefiting both the training of VAEs and NFs.

\begin{wrapfigure}[24]{r}{0.5\linewidth}
    \centering
    \vspace{-3pt} % adjust vertical spacing if needed
  \includegraphics[width=0.45\textwidth,trim=0 10 0 10,clip]{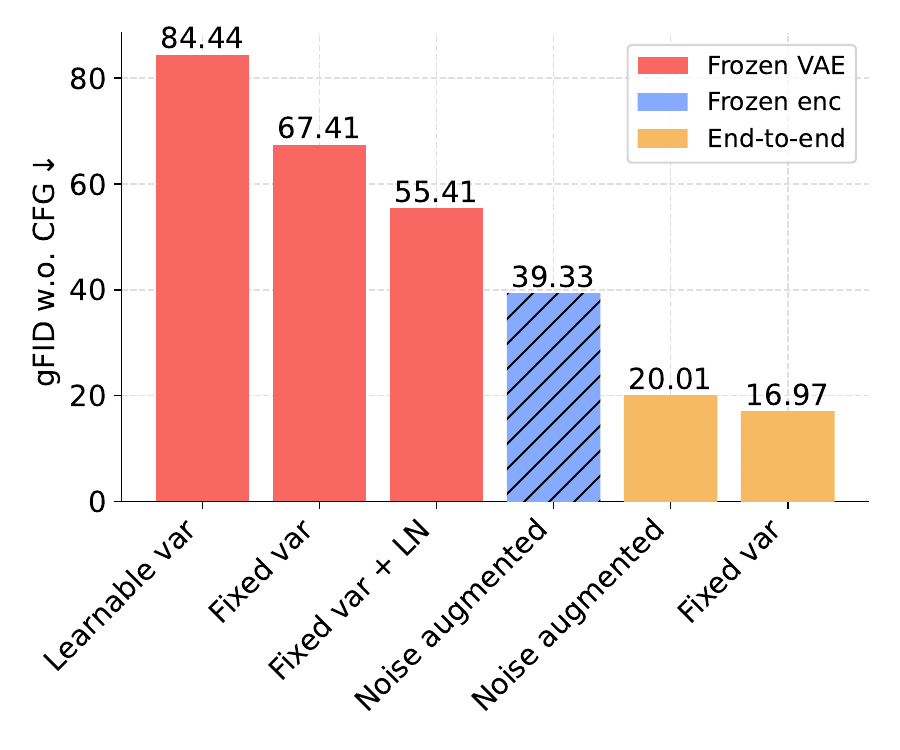}
  %\vskip -0.15in
  \caption{\textbf{Variant studies.} `Frozen VAE' means both VAE encoder and decoder are frozen during training. `Frozen enc' means the decoder is trained. `End-to-end' means VAE encoder and decoder, and the NF are jointly trained from scratch. `Learnable var' means the variance is predicted by the VAE, while  `Fixed var' is our method with $\bar{\sigma}^2=0.5^2$. `LN' denotes applying a layer normalization on the VAE encoder. `Noise augmented' indicates adding Gaussian noise to VAE latents as done by \citet{starflow}.}
  \label{fig:ablation_variant}
% \vskip -0.2in
\end{wrapfigure}

\begin{figure*}
%\vskip -0.15in
  \centering
  \includegraphics[width=0.98\textwidth,trim=5 70 30 0,clip]{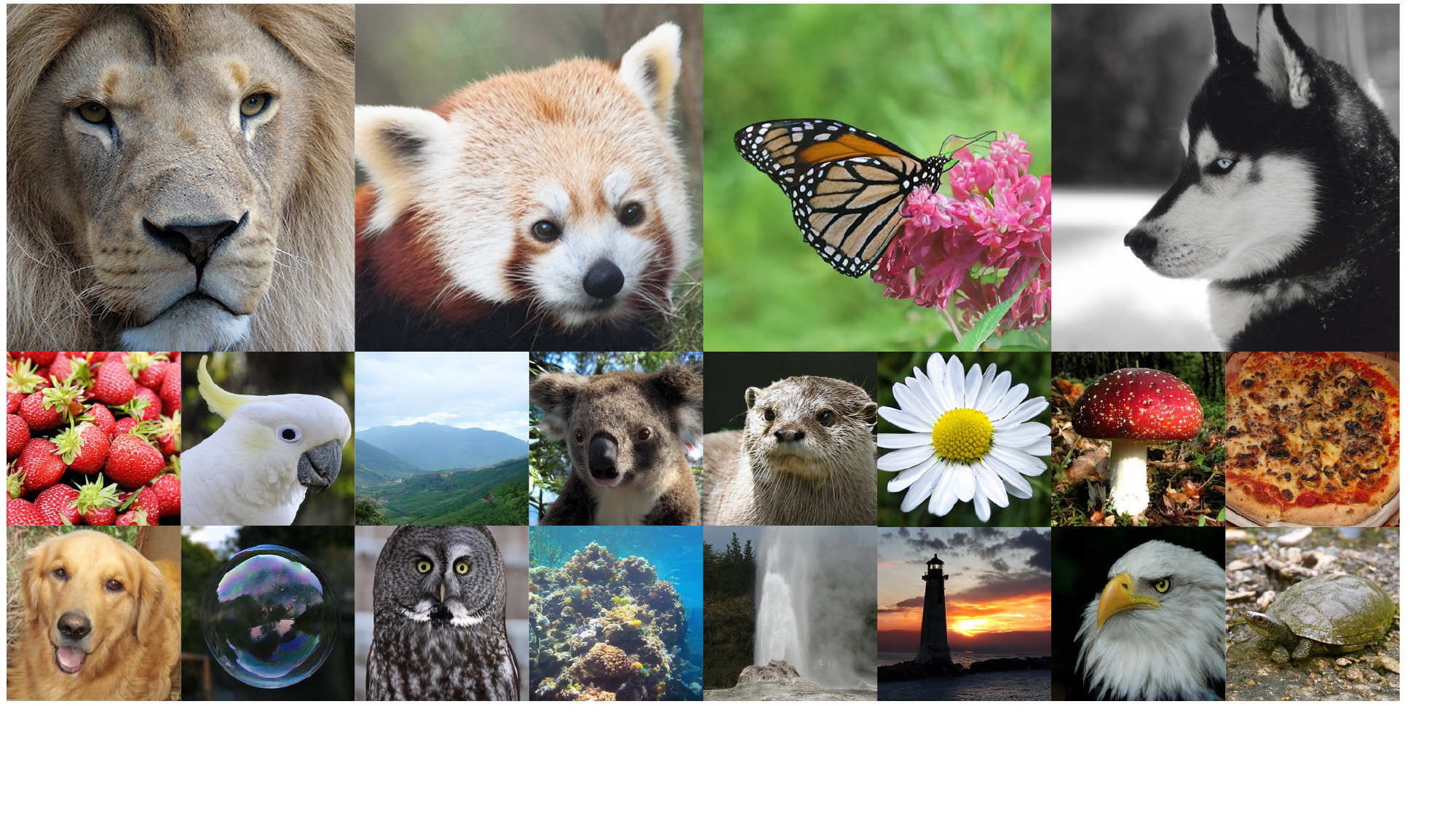}
  %\vskip -0.15in
  \caption{\textbf{Qualitative Results.} We show selected examples generated by SimFlow + REPA-E on ImageNet.}
  \label{fig:generated_samples}
% \vskip -0.1in
\end{figure*}

\textbf{Comparison with state-of-the-art methods on ImageNet 512$\times$512.}
Results are summarized in \tabref{tab:imagenet512_sota}.
SimFlow with REPA-E also works on higher resolution images and significantly surpasses STARFlow. 
Sample images generated from SimFlow are shown in \figref{fig:generated_samples}.

\textbf{Effectiveness of end-to-end training.} We present results in \figref{fig:ablation_variant}. Specifically, we train a normal VAE with fixed variance and then train the NF model without updating the VAE at all, denoted as `Fixed var' under the `Frozen VAE' category. It is clear that our method `Fixed var' under `End-to-end' training significantly outperforms the other (gFID = 16.97 vs. gFID 67.41). Only setting a fixed variance is not enough. The end-to-end training makes the latent space significantly more suitable for generation. Moreover, we apply noise-augmented training~\cite{starflow} with encoder frozen (as in STARFlow) and end-to-end training, respectively. As seen in \figref{fig:ablation_variant}, end-to-end training also obtains lower gFID, indicating its effectiveness.

\textbf{Comparing fixed variance with noise augmentation under end-to-end training.} As shown in \figref{fig:ablation_variant}, fixed variance yields lower gFID (16.97 vs. 20.01). This demonstrates that fixing variance is a useful technique: it simplifies the pipeline while having competitive performance. Here, we do not claim our method is superior to noise augmentation because they essentially share similar principles, \ie, perturbing the latents. We speculate that noise augmentation would have close performance in end-to-end training if we train the model longer or tune the hyperparameters.

\begin{wrapfigure}[16]{r}{0.45\linewidth}
    \centering
    \vspace{-5pt} % adjust vertical spacing if needed
  \includegraphics[width=0.45\textwidth,trim=0 10 0 10,clip]{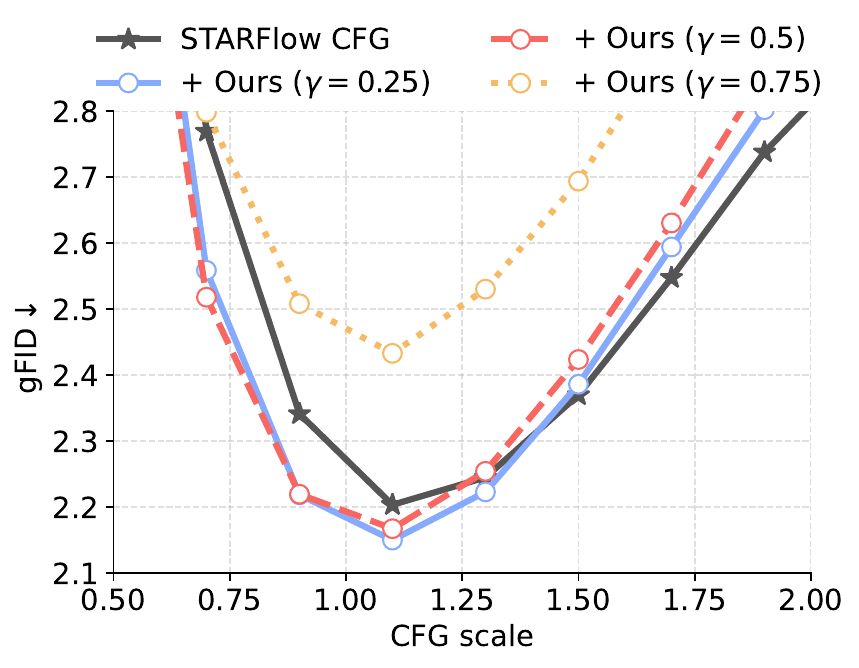}
  %\vskip -0.15in
  \caption{\textbf{CFG method comparison.} Our CFG method in Eq.~\ref{eq:cfg} improves the best point over the STARFlow CFG~\citep{starflow}.}
  \label{fig:cfg}
% \vskip -0.2in
\end{wrapfigure}

\textbf{Reconstruction vs. generation.} The best reconstruction results are achieved by learnable variance (see \secref{sec:supp_variant}). However, \figref{fig:ablation_variant} shows that learnable variance does not necessarily mean better generation. Our observation is consistent with prior works~\citep{vavae,rae}, but from a different perspective of fixing or training the variance term in VAE.

\textbf{Comparing the revised CFG method with STARFlow CFG~\citep{starflow}.} We run STARFlow CFG first to generate tokens and then derive a score-based step, Eq.~\ref{eq:cfg} on the tokens. As demonstrated in \figref{fig:cfg}, with $\gamma=0.25$, our CFG significantly improves the optimal generation quality. We note that our CFG introduces only a mild increase in inference time compared to STARFlow (see \secref{sec:supp_cfg}).

\subsection{Further analysis}\label{sec:further_ana}
\sethlcolor{red!8}
\hl{\textbf{Comparing different latent dimensions.}} Table~\ref{tab:ablation_dim} presents the performance of different VAE latent dimensions. Because we enforce a large and constant variance ($\bar{\sigma}^2 = 0.5^2$), a low latent dimension such as 16 cannot preserve sufficient information. In contrast, a high dimension such as 128, while benefiting reconstruction, would make it challenging for NF modeling. The best trade-off is 64-dimensional. 

\begin{table}[!h]
% \vskip -0.1in
\scriptsize 
    \centering
\caption{\textbf{Comparison of different latent dimensions}. The best trade-off is 64-dimensional. The best results are shown in bold, and the default setting is highlighted in red.}
    \setlength{\tabcolsep}{2.1mm}{
    \begin{tabular}{ccccccc}
        \toprule
        Dim & rFID$\downarrow$ & PSNR$\uparrow$ & LPIPS$\downarrow$ & 
SSIM$\uparrow$ & \makecell{gFID w.o. \\ CFG$\downarrow$} & \makecell{gFID w. \\ CFG$\downarrow$} \\
        \midrule
        16 & 4.98 & 19.94 & 0.327 & 0.45 & \textbf{11.00} & 3.55 \\
        32 & 2.60 & 21.42 & 0.274 & 0.51 & 12.77 & 2.61 \\
        \rowcolor{red!8}64 & 1.49 & 23.01 & 0.222 & 0.59 & 21.44 & \textbf{2.53} \\
        128 & \textbf{0.86} & \textbf{24.61} & \textbf{0.183} & \textbf{0.66} & 33.43 & 3.62 \\
        \bottomrule
    \end{tabular}
}
\label{tab:ablation_dim}
% \vspace{-0.7cm}
\end{table}

\sethlcolor{yellow!10}\hl{\textbf{Comparing different variance values.}} From Table~\ref{tab:ablation_var}, $\bar{\sigma}^2=0.5^2$ generally results in good reconstruction and generation quality. We notice that larger variance accelerates convergence because the latent space is smoother, but reduces the upper bound of the final performance. Overall, the performance of SimFlow is stable across a wide range of variance. 

\begin{table}[!h]
% \vskip -0.1in
\scriptsize 
    \centering
\caption{\textbf{Comparing different levels of variance.} A moderate variance $\bar{\sigma}^2=0.5^2$ performs the best. The superior results are shown in bold, and the default setting is highlighted in yellow. }
    \setlength{\tabcolsep}{2.0mm}{
    \begin{tabular}{lcccccc}
        \toprule
        Var & rFID$\downarrow$ & PSNR$\uparrow$ & LPIPS$\downarrow$ & 
SSIM$\uparrow$ & \makecell{gFID w.o. \\ CFG$\downarrow$} & \makecell{gFID w. \\ CFG$\downarrow$} \\
        \midrule
        $0.1^2$ & 1.90 & 22.61 & 0.238 & 0.57 & 25.54 & 3.01\\
        $0.25^2$ & 1.55 & 22.94 & 0.225 & 0.58 & 21.35 & 2.57 \\
        \rowcolor{yellow!10}$0.5^2$ & \textbf{1.49} & \textbf{23.01} & \textbf{0.222} & \textbf{0.59} & 21.44 & 2.53 \\
        $0.75^2$ & 1.54 & 22.88 & 0.224 & \textbf{0.59} & \textbf{17.76} & \textbf{2.39} \\
        $1.0$ & 1.76 & 22.60 & 0.231 & 0.58  & 19.36 & 2.66 \\
        \bottomrule
    \end{tabular}
}
\label{tab:ablation_var}
% \vspace{-0.7cm}
\end{table}

\sethlcolor{gray!8}\hl{\textbf{Comparing different model sizes.}} In Table~\ref{tab:ablation_nf_size}, scaling up NFs improves
both the reconstruction and generation quality. A small NF cannot fit a complex latent distribution, and thus, the encoder tends to maintain a simple latent space, sacrificing the reconstruction quality. When the NF is larger and more expressive, generation quality is improved. Meanwhile, the encoder can embed more information into latents, enhancing reconstruction quality as well.

\begin{table}[!h]
% \vskip -0.1in
\scriptsize 
    \centering
\caption{\textbf{Comparing different model sizes.} Large NFs improve both the reconstruction and generation quality. The best results are shown in bold, and the default setting is highlighted in gray.}
    \setlength{\tabcolsep}{1.2mm}{
        \begin{tabular}{cccccccc}
        \toprule
        Model & \#Params & rFID$\downarrow$ & PSNR$\uparrow$ & LPIPS$\downarrow$ & 
SSIM$\uparrow$ & \makecell{gFID w.o. \\ CFG$\downarrow$} & \makecell{gFID w. \\ CFG$\downarrow$} \\
        \midrule
        S & 37M & 1.70 & 22.81 & 0.230 & 0.58 & 65.57 & 15.34\\
        B & 141M & 1.66 & 22.89 & 0.226 & 0.58 & 52.77 & 9.72 \\
        L & 475M & 1.52 & 22.95 & 0.224 & \textbf{0.59} & 33.53 & 3.72 \\
        XL & 695M & 1.51 & 22.99 & 0.224 & \textbf{0.59} & 28.60 & 3.51 \\
        \rowcolor{gray!11}XXL & 1.4B & \textbf{1.49} & \textbf{23.01} & \textbf{0.222} & \textbf{0.59} & \textbf{21.44} & \textbf{2.53} \\
        \bottomrule
        \end{tabular}
}
\label{tab:ablation_nf_size}
% \vspace{-0.7cm}
\end{table}

\sethlcolor{green!10}\hl{\textbf{Comparison of different ways of adding noise to latent $\bx$.}} 
We test different latent noise discussed in prior works~\citep{yang2025latent, qiu2025image, meshchaninov2025compressed}. Specifically, we train VAEs with NFs for 100K iterations with these perturbations. As we expect in Section~\ref{sec:working_mech}, \tabref{tab:noise_method} shows that all these methods avoid collapse during end-to-end training. Note that these perturbations have different strengths in removing information in latents, leading to different performance. If we tune the weights of the strengths of these methods, they should lead to similar results, which we leave for future exploration.

\begin{table}[!h]
% \vskip -0.1in
\scriptsize 
    \centering
\caption{\textbf{Comparison of different ways of adding noise to latent $\bx$.} All these ways enable end-to-end training. Our method `Additive' performs the best and is highlighted in green.}
    \setlength{\tabcolsep}{1.7mm}{
\begin{tabular}{lcccc}
\toprule
    Noise & Equation & rFID$\downarrow$  & PSNR$\uparrow$ & gFID w.o. CFG$\downarrow$ \\
    \midrule
No noise & $\bx'=\bx$ & \multicolumn{3}{c}{Collapse} \\
Linear & $\bx'=t\bx + (1-t)\bepsilon$ & 8.54 & 19.74 & 71.0 \\
Slerp & $\bx'=t\bx + \sqrt{1-t^2}\bepsilon$ & 5.56 & 20.63 & 49.7 \\
\rowcolor{green!10}Additive & $\bx'=\bx + \bepsilon$ & 2.90 & 22.22 & 39.6 \\
    \bottomrule
  \end{tabular}
}
\label{tab:noise_method}
% \vspace{-0.7cm}
\end{table}

\sethlcolor{orange!10}\hl{\textbf{Effectiveness of SimFlow methods applied to diffusion models.}}
We perform a preliminary study by jointly training VAE and SiT~\citep{sit}. We choose the VAE in LDM~\citep{sd} and use the 32- and 64-dimensional checkpoints from \citet{vavae}, which were trained for 50 epochs. We compare 1) SiT trained for 50 epochs with the pretrained VAE frozen, and 2) jointly train SiT and a new VAE for 50 epochs from scratch. \tabref{tab:end_to_end_sit} shows our method also speeds up the training of diffusion models and achieves better generation quality after the same iterations. With 80 epochs of end-to-end training, the performance is further improved. End-to-end diffusion model training is a promising future direction.

\begin{table}[!h]
\scriptsize 
    \centering
\caption{\textbf{Applying fixed variance to end-to-end training of diffusion models and VAEs.} The colored rows indicate our methods with fixed variance in VAEs.}
    \setlength{\tabcolsep}{0.7mm}{
\begin{tabular}{ccccccc}
    \toprule
    VAE dim & VAE epochs & Diff. epochs & Joint & rFID$\downarrow$ & PSNR$\uparrow$ & gFID w.o. CFG$\downarrow$ \\
    \midrule
    32 & 50 & 50 & \xmark & 0.26 & 28.59 & 20.36 \\
    \rowcolor{orange!10}32 & 50 & 50 & \cmark & 1.26 & 21.86 & 17.70 (\textcolor{red}{-2.66}) \\
    \arrayrulecolor{black!30}\midrule
    64 & 50 & 50 & \xmark & 0.17 & 31.03 & 28.34 \\
    \rowcolor{orange!10}64 & 50 & 50 & \cmark & 0.82 & 22.83 & 19.10 (\textcolor{red}{-9.24}) \\
    \arrayrulecolor{black!30}\midrule
    \rowcolor{orange!10}64 & 80 & 80 & \cmark & 0.76 & 23.10 & 13.91 \\
    \arrayrulecolor{black}\bottomrule
  \end{tabular}
}
\label{tab:end_to_end_sit}
% \vspace{-0.7cm}
\end{table}

\section{Conclusion}
This paper presents SimFlow, an end-to-end training framework for latent NFs by simply fixing the VAE variance. This makes latent space smoother and helps NFs generalize better when sampling, without needing extra noise schedules or denoising steps. Experiments show that SimFlow improves generation quality and speeds up training compared to existing NF methods. Future work will expand this framework to text-to-image training and explore a second-stage training with the VAE fixed after joint training.

\beginappendix
\section{Training objectives}\label{sec:supp_training_obj}
\subsection{VAE ELBO}
A standard VAE~\citep{kingma2013auto} consists of an encoder $q_{\bpsi}$ and a decoder $p_{\bomega}$. Given an image $\bi$, the encoder predicts mean $\bmu$ and variance $\bsigma^2$ of a Gaussian distribution $\mathcal{N}(\bmu, \text{diag}(\bsigma^2))$. A latent variable $\bx$ (a set of tokens) is sampled from $\mathcal{N}(\bmu, \text{diag}(\bsigma^2))$. The decoder is trained to reconstruct the image $\bi$ from $\bx$. The log likelihood $\log p(\bi)$ for VAEs can be decomposed as follows~\citep{chan2024tutorial}:
\begin{equation}
    \log p(\bi) = \underbrace{\Ex_{q_{\bpsi}(\bx\mid\bi)} \left[ \log \frac{p(\bi, \bx)}{q_{\bpsi}(\bx\mid\bi)} \right]}_{\stackrel{\text{def}}{=}\text{ELBO}(\bi)} + \KL(q_{\bpsi}(\bx\mid\bi)\parallel p(\bx\mid\bi)).\label{eq:log_pi}
\end{equation}
To prove the preceding equation, first, we use the proxy $q_{\bpsi}(\bx\mid\bi)$ to poke around $p(\bi)$.
\begin{align}
    \log p(\bi) &= \log p(\bi) \times \underbrace{\int q_{\bpsi}(\bx\mid\bi)d\bx}_{=1} \quad\quad \text{(multiply 1)} \\
    &= \int \log p(\bi) \times q_{\bpsi}(\bx\mid\bi) d\bx \\
    &= \Ex_{q_{\bpsi}(\bx\mid\bi)}[\log p(\bi)], \label{eq:proxy}
\end{align}
where the last equality is the fact that $\int a \times p_Z(z)dz = \Ex[a] = a$ for any random variable $Z$ and a scalar $a$. Then, we use Bayes theorem, \ie, $p(\bi, \bx) = p(\bx\mid\bi)p(\bi)$:
\begin{align}
    & \Ex_{q_{\bpsi}(\bx\mid\bi)}[\log p(\bi)] \\
    = & \Ex_{q_{\bpsi}(\bx\mid\bi)} \left[ \log \frac{p(\bi, \bx)}{p(\bx\mid\bi)} \right] \quad \text{(Bayes Theorem)} \\
    = & \Ex_{q_{\bpsi}(\bx\mid\bi)} \left[ \log \frac{p(\bi, \bx)}{p(\bx\mid\bi)} \times {\frac{q_{\bpsi}(\bx\mid\bi)}{q_{\bpsi}(\bx\mid\bi)}} \right] \\
    = & \underbrace{\Ex_{q_{\bpsi}(\bx\mid\bi)} \left[ \log \frac{p(\bi, \bx)}{{q_{\bpsi}(\bx\mid\bi)}} \right]}_{\text{ELBO}} + \underbrace{\Ex_{q_{\bpsi}(\bx\mid\bi)} \left[ \log \frac{{q_{\bpsi}(\bx\mid\bi)}}{p(\bx\mid\bi)} \right]}_{\KL(q_{\bpsi}(\bx\mid\bi)\parallel p(\bx\mid\bi))},\label{eq:elbo}
\end{align}
where we recognize that the first term is exactly ELBO, whereas the second term is exactly the KL divergence. Comparing Eq.~\ref{eq:elbo} with Eq.~\ref{eq:proxy}, we get Eq.~\ref{eq:log_pi}. Because the KL divergence is always non-negative, we have
\begin{equation}
    \log p(\bi) \ge \Ex_{q_{\bpsi}(\bx\mid\bi)} \left[ \log \frac{p(\bi, \bx)}{q_{\bpsi}(\bx\mid\bi)} \right].
\end{equation}

Moreover, we can decompose the ELBO as follows:
\begin{align}
    & \text{ELBO}(\bi) \\
    \stackrel{\text{def}}{=} & \Ex_{q_{\bpsi}(\bx\mid\bi)} \left[ \log \frac{p(\bi, \bx)}{q_{\bpsi}(\bx\mid\bi)} \right] \\
    = & \Ex_{q_{\bpsi}(\bx\mid\bi)} \left[ \log \frac{{p(\bi\mid\bx)p(\bx)}}{q_{\bpsi}(\bx\mid\bi)} \right]  \\
    = & \mathbb{E}_{q_{\bpsi}(\bx \mid \bi)} \left[ \underbrace{\log p_{\bomega}(\bi \mid \bx)}_{\text{Reconstruction}} + \underbrace{\log p(\bx)}_{\text{Prior}} - \underbrace{\log q_{\bpsi}(\bx \mid \bi)}_{\text{Entropy}}\right],\label{eq:vae_elbo_last}
\end{align}
where we replaced the inaccessible $p(\bi\mid\bx)$ by its proxy $p_{\bomega}(\bi\mid\bx)$. The prior term is usually chosen as the normal distribution $\mathcal{N}(\zeros, \bI)$ in standard VAEs. In SimFlow, we replace the prior term with the NF-modeled probability (\secref{sec:nf_loss}) and the entropy term simplifies to a constant (\secref{sec:supp_entropy_term}).

\subsection{NF loss}\label{sec:nf_loss}
Latent NFs map the VAE latent distribution into the normal distribution $\bz \sim p_0(\bz)$, via learning an invertible function $f_{\btheta}$. NFs are trained with maximum likelihood estimation, following the change of variable formula:
\begin{align}
& \max_{\btheta}\mathbb{E}_{\bx \sim p_{\text{latent}}} \log p_{\text{NF}}(\bx; {\btheta}). \\
= & \mathbb{E}_{\bx \sim p_{\text{latent}}}\left[ \log p_0(f_{\btheta}(\bx)) + \log \left| \det \left( \frac{\partial f_{\btheta}(\bx)}{\partial \bx} \right) \right| \right] \\
= & \mathbb{E}_{\bx \sim p_{\text{latent}}}\left[ -\frac{1}{2}\lVert f_{\btheta}(\bx) \rVert^2_2 + \log \left| \det \left( \frac{\partial f_{\btheta}(\bx)}{\partial \bx} \right) \right| \right],\label{eq:ll_norm}
\end{align}
where Eq.~\ref{eq:ll_norm} leverages the log-likelihood of the normal distribution and removes terms independent of $\bx$.

We build our study on STARFlow~\citep{starflow} and thus, use the same NF loss with TARFlow~\citep{tarflow}. The forward and inverse processes of a TARFlow block are:
\begin{equation}
    \bz_d = \frac{\bx_d - \bbeta_{\btheta}(\bx_{<d})}{\balpha_{\btheta}(\bx_{<d})},
\bx_d = \bbeta_{\btheta}(\bx_{<d}) + \balpha_{\btheta}(\bx_{<d}) \cdot \bz_d,
\end{equation}
where $d \in [1, D]$ denotes the token id. The Jacobian term in Eq.~\ref{eq:change_of_var} becomes extremely simple:
\begin{equation}
    \log\left| \det \left( \frac{\partial f_{\btheta}(\bx)}{\partial \bx} \right) \right|=-\sum_{d=1}^D\log \balpha_{\btheta}(\bx_{<d}).
\end{equation}
TARFlow~\citep{tarflow} stacks $T$
blocks whose autoregressive ordering alternates from one layer to the next, \eg, in the first layer, from left to right, and in the next layer, from right to left. Training is performed for all blocks:
\begin{equation}
\max_{\btheta}\mathbb{E}_{\bx \sim p_{\text{latent}}}\left[ -\frac{1}{2}\lVert f_{\btheta}(\bx) \rVert^2_2 -\sum_{t=1}^T \sum_{d=1}^D\log \balpha^t_{\btheta}(\bx^t_{<d}) \right].
\label{eq:nf_last}
\end{equation}

\subsection{Constant entropy term in SimFlow}\label{sec:supp_entropy_term}
Now, we analyze the entropy term in Eq.~\ref{eq:vae_elbo_last}.
\begin{align}
    & -\mathbb{E}_{q_{\bpsi}(\bx \mid \bi)} \left[ \log q_{\bpsi}(\bx \mid \bi) \right] \\
    = & -\mathbb{E}_{q_{\bpsi}(\bx \mid \bi)} \left[\log \prod_{i=1}^N \frac{1}{ \sigma_i \sqrt{2\pi}} 
       \exp\!\left(-\frac{1}{2} \left(\frac{x_i - \mu_i}{\sigma_i}\right)^{2}\right)\right] \\
= & -\mathbb{E}_{q_{\bpsi}(\bx \mid \bi)} \left[-\sum_{i=1}^N \left(\log ( \sigma_i \sqrt{2\pi}) + 
       \frac{1}{2} \left(\frac{x_i - \mu_i}{\sigma_i}\right)^{2}\right)\right] \\
= & \frac{1}{2}\sum_{i=1}^N \log ( 2\pi\sigma_i^2 ) + 
\frac{1}{2}\mathbb{E}_{q_{\bpsi}(\bx \mid \bi)} \left[
       \sum_{i=1}^N \left(\frac{x_i - \mu_i}{\sigma_i}\right)^{2}\right].
\end{align}
Note that
\begin{align}
& \mathbb{E}_{q_{\bpsi}(\bx \mid \bi)} \left[
       \sum_{i=1}^N \left(\frac{x_i - \mu_i}{\sigma_i}\right)^{2}\right] \\
= & \sum_{i=1}^N \mathbb{E}_{x_i\sim \mathcal{N}(\mu_i, \sigma_i^2)} \left[
        \left(\frac{x_i - \mu_i}{\sigma_i}\right)^{2}\right]\\
= & \sum_{i=1}^N \mathbb{E}_{x_i\sim \mathcal{N}(\mu_i, \sigma_i^2)} \left[
        \frac{(x_i - \mu_i)^2}{\sigma_i^2}\right]\\
        = & N \quad (\text{Var}(x) = \mathbb{E}[(x-\mu)^2]).
\end{align}
Thus,
\begin{equation}
-\mathbb{E}_{q_{\bpsi}(\bx \mid \bi)} \left[ \log q_{\bpsi}(\bx \mid \bi) \right] = \frac{1}{2}\sum_{i=1}^N \log ( 2\pi\sigma_i^2 ) + 
\frac{1}{2}N.\label{eq:entropy_last}
\end{equation}
Because we set all $\sigma_i$ to be a fixed value, the entropy term also becomes a constant in SimFlow.

\subsection{Objective of end-to-end training}
Last, we combine Eq.\ref{eq:vae_elbo_last}, Eq.\ref{eq:nf_last}, with Eq.~\ref{eq:entropy_last} and we omit the constant entropy term. The objective of our end-to-end training is presented as follows:
\begin{align}
    & \max_{\btheta,\bpsi,\bomega}\mathbb{E}_{\bi \sim p_{\text{data}}} \mathbb{E}_{\bx\sim q_{\bpsi}(\bx \mid \bi)} \left[ \log p_{\bomega}(\bi \mid \bx) + \log p_{\text{NF}}(\bx; {\btheta}) \right] \\
    = & \mathbb{E}_{\bi \sim p_{\text{data}}} \mathbb{E}_{\bx\sim q_{\bpsi}(\bx \mid \bi)} \left[ \log p_{\bomega}(\bi \mid \bx) - \frac{1}{2}\lVert f_{\btheta}(\bx) \rVert^2_2 - \sum_{t=1}^T \sum_{d=1}^D\log \balpha^t_{\btheta}(\bx^t_{<d}) \right].
\end{align}
Note that the reconstruction loss (the first term) is typically averaged at the pixel level, while the NF loss terms (the second and third terms) are normalized at the latent feature level. Therefore, in the specific implementation of our codes, we average the loss terms at different levels. We do not further tune the weight of each term. Besides, we also add perceptual loss and adversarial loss~\citep{gan} for VAE training~\citep{sd}.

\section{Implementation details}\label{sec:supp_imple_detail}
\textbf{VAE.} We employ the MAR VAE architecture \citep{mar}, which is adapted from the LDM VAE \citep{sd} and comprises approximately 67 million parameters. We maintain most of the original hyperparameters from \citet{mar}, modifying only the latent dimension (\eg, 16, 32, or 64). The patch size remains fixed at 16. Additionally, following recent findings that highlight the benefits of normalization layers in VAE encoders \citep{rae}, we also apply a Layer Normalization to each token independently. This normalizes the tokens into a constrained space and improves the gFID score by about 0.15.

\begin{table}[b]
% \vskip -0.1in
\scriptsize 
    \centering
\caption{\textbf{Model configurations for different sizes.} Depth `2, 2, 2, 2, 2, x' denote that each of the first five blocks consists of only two layers while the last block contains `x' layers.}
    \setlength{\tabcolsep}{2.5mm}{
    \begin{tabular}{lcccc}
        \toprule
        Model & \#Params &  Dim & Num-Heads & Depth \\
    \midrule
    S & 37M & 384 & 6 & 2, 2, 2, 2, 2, 2 \\
    B & 141M & 768 & 12 & 2, 2, 2, 2, 2, 2 \\
    L & 475M & 1024 & 16 & 2, 2, 2, 2, 2, 14 \\
    XL & 695M & 1152 & 16 & 2, 2, 2, 2, 2, 18 \\
    XXL & 1.4B & 1152 & 16 & 2, 2, 2, 2, 2, 36 \\
    \bottomrule
    \end{tabular}
}
\label{tab:model_conf}
% \vspace{-0.7cm}
\end{table}

\textbf{NF.} We utilize TARFlow \citep{tarflow} as the foundational NF model, because it is the latest open-source NF model. We implement two upgrades to TARFlow. First, we adopt a deep-shallow architecture, following \citet{starflow}, where the final block contains multiple layers while all other blocks consist of only two layers. Second, we incorporate the adaLN-Zero mechanism into the NF for class conditioning, following the design of DiT \citep{dit} and SiT \citep{sit}. A detailed configuration of our models is provided in Table~\ref{tab:model_conf} and the training configuration is presented in Table~\ref{tab:train_conf}.

\begin{table}
% \vskip -0.1in
\scriptsize 
    \centering
\caption{\textbf{Training configurations for SimFlow.}}
    \setlength{\tabcolsep}{2.5mm}{
    \begin{tabular}{lc}
        \toprule
Warm-up epochs & 0 \\
Pretraining epochs & 0 \\
\rowcolor{lightgray} \textbf{Stage-I} & \\
Epochs & 80 \\
Learning rate & $1.0\times 10^{-4}$ \\
Learning rate schedule & constant \\
\rowcolor{lightgray} \textbf{Stage-II} & \\
Epochs & 80 \\
Learning rate & $1.0\times 10^{-4}\rightarrow 1.0\times 10^{-6}$ \\
Learning rate schedule & cosine rate \\
\rowcolor{lightgray} \textbf{Stage-I and -II} & \\
Image size & 256 or 512 \\
Optimizer & AdamW~\citep{adamw}, $\beta_1, \beta_2 = 0.9, 0.999$ \\
Batch size & 256 \\
Weight decay & 0 \\
EMA rate & 0.9999 \\
Max gradient norm & 1.0 \\
Class token drop (for CFG) & 0.1 \\
\rowcolor{lightgray} \textbf{REPA-E} & \\
REPA loss coef & 1.0 \\
Align depth & layer 1 at the block 3 \\
Encoder & DINOv2-B~\citep{dino} \\
    \bottomrule
    \end{tabular}
}
\label{tab:train_conf}
% \vspace{-0.7cm}
\end{table}

\begin{figure*}
%\vskip -0.15in
  \centering
\includegraphics[width=0.98\textwidth,trim=0 300 0 0,clip]{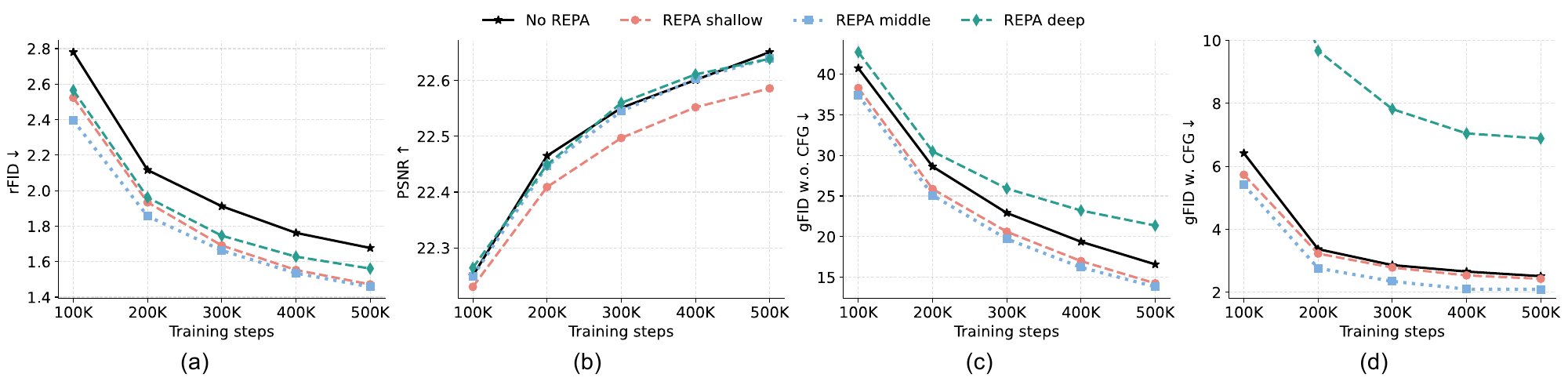}
  %\vskip -0.15in
  \caption{\textbf{Effects of REPA-E on SimFlow.} Aligning a middle block leads to the most consistent gains.}
  \label{fig:repa}
% \vskip -0.2in
\end{figure*}

\textbf{REPA-E.} During training, we use a three-layer multilayer perceptron (MLP) to align the hidden features from an NF block with features extracted by a frozen DINOv2-B model~\citep{dino}. The MLP itself is trainable. As detailed in Section \ref{sec:repa_block}, we compare three different choices for the NF block to align and find that aligning a middle block within the NF model yields the best performance.

\section{Additional experiments}
\subsection{Variant studies}\label{sec:supp_variant}
\tabref{tab:variant_detailed} presents detailed results of variants, which is partly visualized in \figref{fig:ablation_variant}. The best reconstruction results are achieved by learnable variance (rFID 0.23) but it does not necessarily mean better generation (gFID 84.44). In comparison, while end-to-end training slightly degrades the reconstruction, the generation quality is significantly improved. Our observation is consistent with prior works~\citep{vavae,rae}, but from a different perspective of fixing or training the variance term in VAEs.

\begin{table}[t]
% \vskip -0.1in
\scriptsize 
    \centering
\caption{\textbf{Variant studies.} `Frozen VAE' means both VAE encoder and decoder are frozen during training. `Frozen enc' means the decoder is trained. `End-to-end' means VAE encoder and decoder, and the NF are jointly trained from scratch. `Learnable var' means the variance is predicted by the VAE, while  `Fixed var' is our method with $\bar{\sigma}^2=0.5^2$. `LN' denotes applying a layer normalization on the VAE
encoder. `Noise' indicates adding Gaussian noise to VAE latents as done by \citet{starflow}.}
    \setlength{\tabcolsep}{2.0mm}{
    \begin{tabular}{lcccc}
        \toprule
        Config & Method & rFID$\downarrow$ & PSNR$\uparrow$ & \makecell{gFID w.o. \\ CFG$\downarrow$} \\
        \midrule
Frozen VAE & Learnable var & 0.23 & 29.76 &  84.44 \\
& Fixed var & 0.40 & 27.54 & 67.41 \\
& Fixed var + LN & 1.07 & 24.43 & 55.41 \\
\arrayrulecolor{black!30}\midrule
Frozen Encoder & Noise & 4.07 & 19.94 & 39.33 \\
\arrayrulecolor{black!30}\midrule
End-to-end & Noise & 1.39 & 23.20 & 20.01 \\
& Fixed var & 1.76 & 22.61 & 16.97 \\
\arrayrulecolor{black}\bottomrule
    \end{tabular}
}
\label{tab:variant_detailed}
% \vspace{-0.7cm}
\end{table}

\subsection{Choice of NF blocks for REPA-E}\label{sec:repa_block}
We align different blocks of NFs: the first block, a middle block, or the last one. The results are shown in \figref{fig:repa}. As seen, introducing alignment at shallow layers negatively affects the VAE reconstruction performance (lower PSNR), whereas applying alignment at deeper layers harms the generation quality. The reason is that aligning the first block mainly affects the VAE latents and only has a slight impact on NFs. Moreover, in the last block, we expect that data distribution is close to the noise distribution, so this block is not suitable for semantic understanding and alignment. As a result, aligning a middle block within the NF model leads to the most consistent gains.

\subsection{Efficiency of our CFG method}\label{sec:supp_cfg}
\begin{wrapfigure}[9]{r}{0.2\textwidth}
    \centering
    \vspace{-0.25in}
    % \hspace{-1.2in}
    \includegraphics[width=\linewidth]{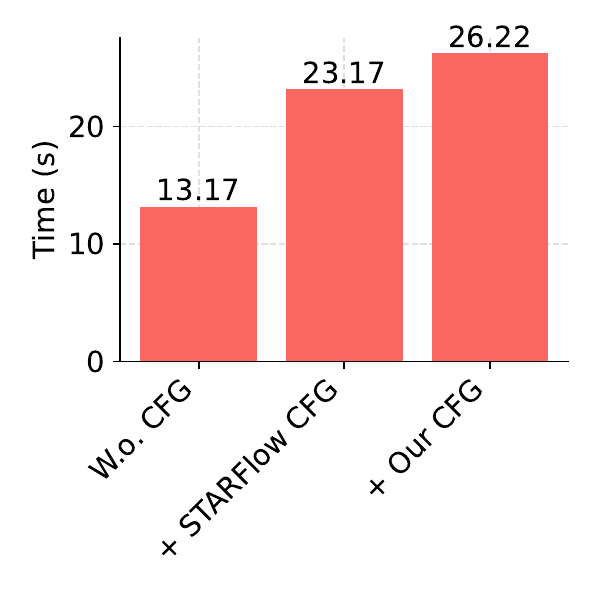}
    % \hspace{-0.8in}
    \label{fig:t_cfg}
\end{wrapfigure}
We measure the average inference time of generating a batch of 256 images for SimFlow 1) without CFG, 2) with STARFlow CFG, and 3) with STARFlow + our CFG. As shown in the right figure, our method only mildly increases the inference time compared to the STARFlow CFG baseline.

\subsection{Qualitative Results}
Figures~\ref{fig:results1_repae} to \ref{fig:results4_repae} present additional uncurated examples on ImageNet 256×256 with SimFlow+REPA-E. We follow the practice of \citet{jit}.

\newcommand{\hhs}{\hspace{-0.001em}}
\newcommand{\vvs}{\vspace{-.1em}}

\newcommand{\tilewidth}{0.14\linewidth}

\newcommand{\imgcaptiontext}{
\textit{Uncurated} samples on ImageNet 256$\times$256 using SimFlow + REPA-E conditioned on the specified classes. Unlike the common practice of visualizing with a higher CFG, here we show images using the CFG value that achieves the reported gFID of 1.91.}

\newcommand{\sampledir}{samples256_simflow_repae_jpg}

\newcommand{\addclass}[2]{
\begin{minipage}[t]{0.49\linewidth}
\centering
\includegraphics[width=\tilewidth]{\sampledir/cls#1/000#1.jpg}\hhs
\includegraphics[width=\tilewidth]{\sampledir/cls#1/001#1.jpg}\hhs
\includegraphics[width=\tilewidth]{\sampledir/cls#1/002#1.jpg}\hhs
\includegraphics[width=\tilewidth]{\sampledir/cls#1/003#1.jpg}\hhs
\includegraphics[width=\tilewidth]{\sampledir/cls#1/004#1.jpg}\hhs
\includegraphics[width=\tilewidth]{\sampledir/cls#1/005#1.jpg}\hhs
\includegraphics[width=\tilewidth]{\sampledir/cls#1/006#1.jpg}\vvs
\\
\includegraphics[width=\tilewidth]{\sampledir/cls#1/007#1.jpg}\hhs
\includegraphics[width=\tilewidth]{\sampledir/cls#1/008#1.jpg}\hhs
\includegraphics[width=\tilewidth]{\sampledir/cls#1/009#1.jpg}\hhs
\includegraphics[width=\tilewidth]{\sampledir/cls#1/010#1.jpg}\hhs
\includegraphics[width=\tilewidth]{\sampledir/cls#1/011#1.jpg}\hhs
\includegraphics[width=\tilewidth]{\sampledir/cls#1/012#1.jpg}\hhs
\includegraphics[width=\tilewidth]{\sampledir/cls#1/013#1.jpg}\vvs
\\
\includegraphics[width=\tilewidth]{\sampledir/cls#1/014#1.jpg}\hhs
\includegraphics[width=\tilewidth]{\sampledir/cls#1/015#1.jpg}\hhs
\includegraphics[width=\tilewidth]{\sampledir/cls#1/016#1.jpg}\hhs
\includegraphics[width=\tilewidth]{\sampledir/cls#1/017#1.jpg}\hhs
\includegraphics[width=\tilewidth]{\sampledir/cls#1/018#1.jpg}\hhs
\includegraphics[width=\tilewidth]{\sampledir/cls#1/019#1.jpg}\hhs
\includegraphics[width=\tilewidth]{\sampledir/cls#1/020#1.jpg}\vvs
\\
{\scriptsize {class #1}: #2}
\vspace{.5em}
\end{minipage}
}

\begin{figure*}[t]
\centering
\addclass{012}{house finch, linnet, Carpodacus mexicanus}
\addclass{014}{indigo bunting, indigo finch, indigo bird, Passerina cyanea
}
\\
\addclass{042}{agama}
\addclass{081}{ptarmigan}
\\
\addclass{107}{jellyfish}
\addclass{108}{sea anemone, anemone}
\\
\addclass{110}{flatworm, platyhelminth}
\addclass{117}{chambered nautilus, pearly nautilus, nautilus}
\\
\addclass{130}{flamingo}
\addclass{279}{Arctic fox, white fox, Alopex lagopus}
\\
\caption{\imgcaptiontext}
\label{fig:results1_repae}
\vspace{-1em}
\end{figure*}

\begin{figure*}[t]
\centering
\addclass{288}{leopard, Panthera pardus}
\addclass{309}{bee}
\\
\addclass{349}{bighorn, bighorn sheep, cimarron, Rocky Mountain bighorn}
\addclass{397}{puffer, pufferfish, blowfish, globefish}
\\
\addclass{425}{barn}
\addclass{448}{birdhouse}
\\
\addclass{453}{bookcase}
\addclass{458}{brass, memorial tablet, plaque}
\\
\addclass{495}{china cabinet, china closet}
\addclass{500}{cliff dwelling}
\\
\caption{\imgcaptiontext}
\label{fig:results2_repae}
\vspace{-1em}
\end{figure*}

\begin{figure*}[t]
\centering
\addclass{658}{mitten}
\addclass{661}{Model T}
\\
\addclass{718}{pier}
\addclass{724}{pirate, pirate ship}
\\
\addclass{725}{pitcher, ewer}
\addclass{757}{recreational vehicle, RV, R.V.}
\\
\addclass{779}{school bus}
\addclass{780}{schooner}
\\
\addclass{829}{streetcar, tram, tramcar, trolley, trolley car}
\addclass{853}{thatch, thatched roof}
\\
\caption{\imgcaptiontext}
\label{fig:results3_repae}
\vspace{-1em}
\end{figure*}

\begin{figure*}[t]
\centering
\addclass{873}{triumphal arch}
\addclass{900}{water tower}
\\
\addclass{911}{wool, woolen, woollen}
\addclass{913}{wreck}
\\
\addclass{927}{trifle}
\addclass{930}{French loaf}
\\
\addclass{946}{cardoon}
\addclass{947}{mushroom}
\\
\addclass{975}{lakeside, lakeshore}
\addclass{989}{hip, rose hip, rosehip}
\\
\caption{\imgcaptiontext}
\label{fig:results4_repae}
\vspace{-1em}
\end{figure*}

\clearpage

\bibliographystyle{plainnat}
\bibliography{main}

\end{document}